\documentclass[10pt,journal,compsoc]{IEEEtran}
\usepackage{times}
\usepackage{graphicx,subfigure}
\usepackage{lscape}

\usepackage{graphicx}
\usepackage{amsmath}
\usepackage{amssymb}
\usepackage{amsthm}
\usepackage{cite}
\usepackage{url}
\usepackage{color}
\usepackage{verbatim}
\definecolor{Red}{rgb}{1, 0.2, 0.2}

\ifCLASSINFOpdf

\else

\fi

\begin{document}

\title{Fine-Grained Classification of Cervical Cells Using Morphological and Appearance Based Convolutional Neural Networks}

\author{\IEEEauthorblockN{Haoming~Lin,
        Yuyang~Hu,
		Siping~Chen,
		Jianhua~Yao,
        and~Ling~Zhang}
\thanks{This work was supported by the National Natural Science Foundation of China (Grant No. 81601510, 81501545), National Natural Science Foundation of Guangdong Province (Grant No.2016A030310047), New teacher natural science research project of Shenzhen University (Grant No.2018011). 
\emph{Asterisk indicates corresponding authors.}}
\thanks{H. Lin*, Y. Hu, and S. Chen are with the Shenzhen University School of Medicine, Shenzhen University, Shenzhen, China; National-Regional Key Technology Engineering Laboratory for Medical Ultrasound, Shenzhen, China; Guangdong Key Laboratory for Biomedical Measurements and Ultrasound Imaging, Shenzhen, China. e-mail: (hm\_lin@szu.edu.cn).}
\thanks{J. Yao is with the Tencent Holdings Limited, Shenzhen 518057, China.}
\thanks{L. Zhang* is with the Nvidia Corporation, Bethesda, MD 20814, USA. e-mail: (zhangling0722@163.com).}
\thanks{This work was performed when J. Yao and L. Zhang were in the National Institutes of Health.}
}

\maketitle

\begin{abstract}
    Fine-grained classification of cervical cells into different abnormality levels is of great clinical importance but remains very challenging. Contrary to traditional classification methods that rely on hand-crafted or engineered features, convolution neural network (CNN) can classify cervical cells based on automatically learned deep features. However, CNN in previous studies do not involve cell morphological information, and it is unknown whether morphological features can be directly modeled by CNN to classify cervical cells. This paper presents a CNN-based method that combines cell image appearance with cell morphology for classification of cervical cells in Pap smear. The training cervical cell dataset consists of adaptively re-sampled image patches coarsely centered on the nuclei. Several CNN models (AlexNet, GoogleNet, ResNet and DenseNet) pre-trained on ImageNet dataset were fine-tuned on the cervical dataset for comparison. The proposed method is evaluated on the Herlev cervical dataset by five-fold cross-validation at patient level splitting. Results show that by adding cytoplasm and nucleus masks as raw morphological information into appearance-based CNN learning, higher classification accuracies can be achieved in general. Among the four CNN models, GoogleNet fed with both morphological and appearance information obtains the highest classification accuracies of 94.5\% for 2-class classification task and 64.5\% for 7-class classification task. Our method demonstrates that combining cervical cell morphology with appearance information can provide improved classification performance, which is clinically important for early diagnosis of cervical dysplastic changes.
\end{abstract}

\begin{IEEEkeywords}
    Fine-grained classification, Cell morphology, Deep learning, Pap smear.
\end{IEEEkeywords}

\IEEEpeerreviewmaketitle

\section{Introduction}

\IEEEPARstart{C}{ervical} cancer is one of the most common lethal malignant disease among woman \cite{Jemal2011Global}. The greatest factor for cervical cancer is the infection with some types of human papilloma virus (HPV) which may lead to dysplastic changes in cells before development of cervical cancer \cite{Gadducci2011smoking}. These dysplastic changes of cells typically develop over a prolonged process and refer to a wide spectrum of abnormality. Pap smear, one of the most popular screening tests for prevention and early detection of cervical cancer, has been extensively used in developed countries and  credited with reducing the mortality rate of cervical cancer significantly \cite{saslow2012american}. However, population-wide screening is still not widely available in developing countries \cite{saslow2012american}, partly due to the tedious and complexity nature of manually screening of the abnormal cells from a cervical cytology specimen \cite{birdsong1996automated}. Such diagnosis is also subject to error even for experienced doctors \cite{birdsong1996automated}. To address these concerns, automation-assisted reading systems have been developed to improve efficiency and increase availability over the past few decades. These automation-assisted reading systems are based on automated image analysis techniques \cite{birdsong1996automated,zhang2014automation,bengtsson2014screening}, which automatically select potentially abnormal cells from a given cervical cytology slide for further review and fine-grained classification by the cytoscreener or cytopathologist. According to the World Health Organization classification system, premalignant dysplastic changes of cells can include four stages, which are mild, moderate, severe dysplasia and carcinoma in situ (CIS) \cite{demay2007practical}. The lesions are generally no more than manifestations of HPV infection for the mild stage, but the risk of progression to cancer is relatively high for the more severe stages if not detected and treated \cite{canavan2000cervical}. Early stage of dysplastic changes is important for preventing the developments of precancerous cells. Therefore, fine-grained classification of cervical cells into different stages is of great importance for early stage of dysplastic changes, but works in previous studies based on automated image analysis mainly focus on classification task of normal and abnormal cervical cells \cite{birdsong1996automated,bengtsson2014screening,gao2016hep,zhang2014segmentation,chankong2014automatic,plissiti2012overlapping,guo2012discriminative,nanni2010local}.

The classification process of cervical cell images usually requires that single cells in the slides can be isolated and analyzed. Thus, automation-assisted reading system generally comprises three steps: cell segmentation, feature extraction/selection, and cell classification. When dysplastic changes happen, cervical cells undergo various morphological changes which include changes in terms of size, shape, intensity and texture. Thus, feature descriptors are designed to describe these changes by researchers. The feature design and selection are one of the most important factors for cervical cell classification.

Previously, the extracted features can be grouped into handcrafted features, engineering features and their combination. Handcrafted features \cite{chankong2014automatic,marinakis2008particle,jantzen2005pap} describe the morphology and chromatin characteristics according to ``The Bethesda System (TBS)", and engineering features (13,14) represent texture distribution according to the experience in designing computer-assist-diagnosis algorithms. The resulting features are then organized by using a feature selection or dimensionality reduction algorithm and feeding into the classifier. However, Handcrafted features are hindered by limited understanding of cervical cytology. Engineering features are derived from an unsupervised manner, and thus encode redundant information. The feature selection process potentially ignores significant clues and removes complementary information. 

In addition to feature descriptors, classifiers also play an important role in classification performance. In the previous studies, the linear discriminant analysis \cite{van2002automated}, maximum likelihood \cite{tucker1998Interval}, support vector machines \cite{plissiti2011combining}, k-Nearest Neighbors \cite{Jusman2014Intelligent} and decision trees \cite{Jusman2014Intelligent} are commonly used for cervical cells classification. As single classifier may not be perfect, multi classifiers are integrated by weighted majority voting to obtain a final decision in cervical cells classification task \cite{bora2017automated}.

In the past few years, deep convolutional neural networks (CNN) have been shown to be great success in many computer vision tasks when training on large-scale annotated datasets (i.e. ImageNet) \cite{lecun2015deep}. In contrast to classical machine learning methods that use a series of handcrafted features, CNNs automatically learn multi-level features from the training data set. As the development of more powerful hardware with higher computing power (i.e., Graphics Processing Units, GPUs), CNN architecture has become more and more deep and complicated. A variety of CNN models have been introduced in the literatures, such as LeNet \cite{lecun1989backpropagation}, AlexNet \cite{krizhevsky2012imagenet}, GoogleNet \cite{Szegedy2015going}, ResNet \cite{He2016Deep}, DenseNet \cite{Huang2017Densely} and their variants and so on. The original LeNet only consists of 5 layers while the ResNet has already surpassed 100 layers, even reach to more than 1000 layers. In addition to increase depth directly, the GoogleNet introduces an interception module, which concatenates feature-maps produced by filters of different sizes, to make the network wider and deeper. ResNets have achieved state-of-the-art performance on many challenging image recognition, localization, and detection task, such as ImageNet object detection. Large amounts of labelled data are crucial to the performance of CNN. However, the labelled data is limited for cervical cells images because high quality annotation is costly and challenging even if for experts. Fortunately, transfer learning \cite{yosinski2014transferable} is an effective method to address this problem. CNNs have already significantly improved performance in various medical imaging analysis applications, such as classification of lung diseases in CT \cite{shin2016deep} and X-ray images \cite{wang2017chestx}, segmentation of retinal tissues in OCT images \cite{de2018clinically}, detection of lesions in abdominal CT \cite{yan2018deep}, and prediction of pancreas tumor growth in PET/CT \cite{zhang2018convolutional}. But it is still unclear which network or what is the best network depth and width for cervical cells classification given limited training data.

Besides being directly used as a classifier, CNNs can be used as feature selectors. The basic components of a CNN are stacks of different types of specialized layers, such as convolutional, activation, pooling, and fully connected layer. When training with large-scale data, low-to-high-level features of data can be obtained from the shallow convolutional layer to the bottomless convolutional layer of CNN. It is noted that features learned from the first convolutional layer are similar to Gabor filters and color blobs which can be transferred to other image dataset or task \cite{yosinski2014transferable}. Moreover, learned features extracted from pre-trained model can also be combined with existing handcrafted features, such as local binary pattern (LBP), Histogram of Oriented Gradient (HOG), and then fed to other classifier (e.g. support vector machine, SVM) \cite{NANNI2018ensemble}. However, it is still unknown whether morphological features can be directly modeled by CNN to classify cervical cells. Although CNN has recently been used to classify cervical cells \cite{Zhang2017DeepPap}, several problems need more investigation, such as: 1) CNN in previous study does not involve cell morphological information; 2) only evaluate AlexNet which may not represent the capability of the state-of-art deep classification models; 3) previous methods with five-fold cross-validation (CV) does not guarantee patient-level separation on the Herlev dataset, which does not meet the real clinical setting.

In this paper, we present a CNN-based method that combines cell image appearance with cell morphology for classification of cervical cells in Pap smear. In our approach, cell morphology was directly represented by cytoplasm and nucleus binary masks, which were then combined with raw RGB-channels of the cell image to form a five-channel image, on which training data was sampled from a square image patch coarsely centered on the nucleus. Then this dataset was fed into CNNs for classification of cervical cell. Different CNN models (AlexNet, GoogleNet, ResNet and DenseNet) pre-trained on ImageNet dataset were fine-tuned for 2-class and 7-class cervical cell classification based on deep hierarchical features. Herlev dataset consisting of 917 cervical cell images is used to test our method. We carefully split the cells to perform five-fold cross-validation at patient-level -- this means that all the cells from the same patient will be assigned to training set or validation set alone.  Experimental results demonstrated that by adding cytoplasm and nucleus masks as raw morphological information into conventional appearance-based CNN learning, higher classification accuracies can be achieved in general. Among the four CNN models, GoogleNet fed with both morphological and appearance information obtains the highest classification accuracies of 94.5\% for 2-class classification task and 64.5\% for 7-class classification task.

Our main contributions are summarized as follows: 1) the combination of raw cytoplasm and nucleus binary masks and RGB appearance was proposed for CNN-based cervical cell classification. 2) State-of-the-art CNN models were fine-tuned to evaluate and compare the performances of cervical cell classification at patient-level cell splitting. 3) Besides distinguishing normal and abnormal cervical cells, the performances of 7-class fine-grained classification of cervical cell were also investigated.

\section{Methods}
In this study, the cervical cell images which concatenates cytoplasm/nucleus binary masks and raw RGB channels were fed into CNNs, and both the 2-class and 7-class classification performances of state-of-the-art CNN models were evaluated and analyzed. The details are described as below. 

   \begin{figure*}[!t]
   \begin{tabular}{c}
   \includegraphics[width=17cm]{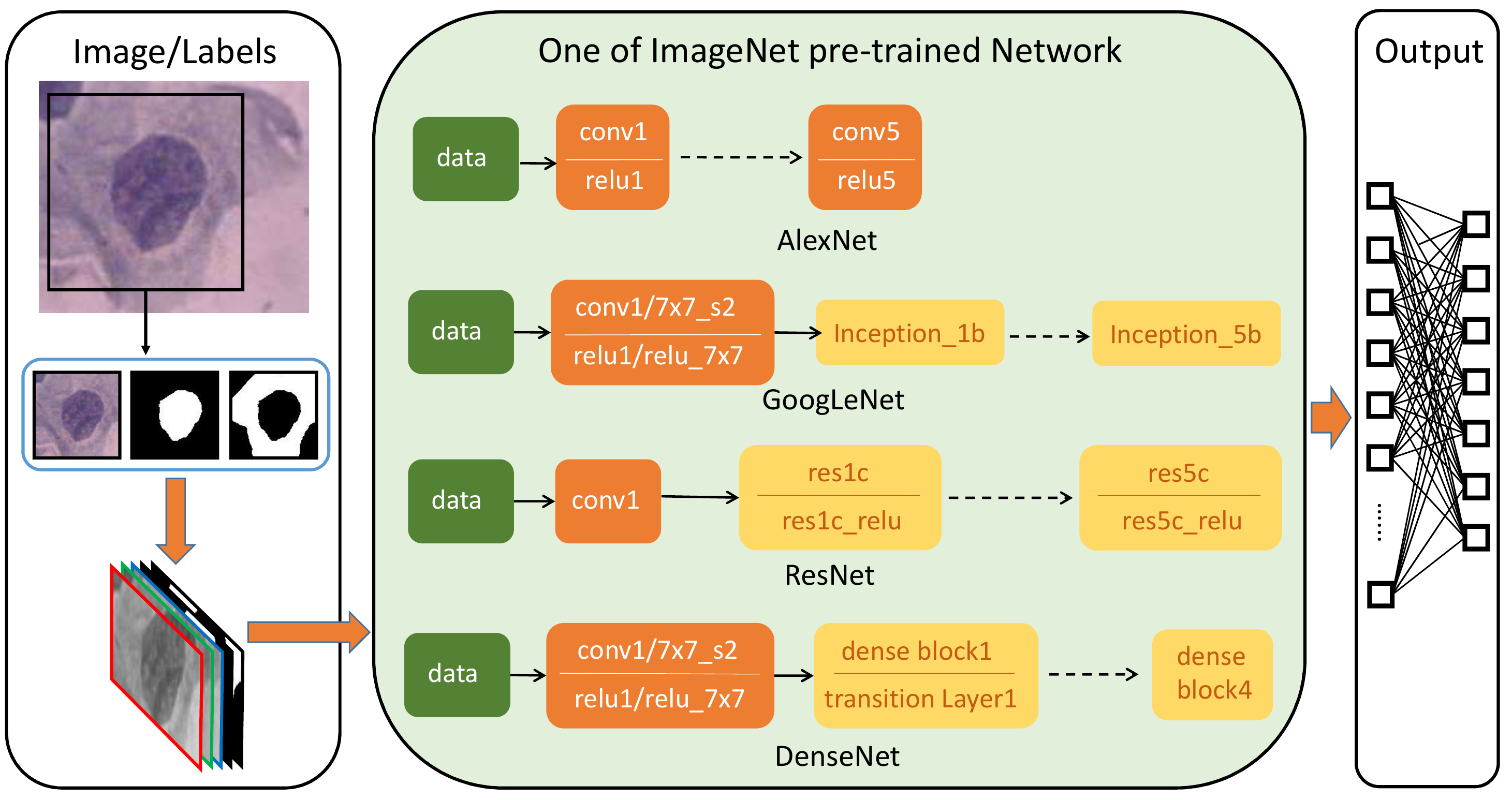}
   \end{tabular}
   \caption[example] 
   { \label{fig_method}
   The overall flow-chart of our CNN framework for 7-class classification problem
}
   \end{figure*} 

\subsection{Data preprocssing}
As mention in the TBS rules, different stages of cervical cytology abnormalities are associated with different nucleus characteristics. Therefore nucleus features in themselves already include substantial discriminative information. Since the main topic of this work is CNN classification, we follow the strategy in \cite{Zhang2017DeepPap} to extract training samples. Specifically, image patches of size $m \times m$ centered on the nucleus centroid and included a certain amount of cytoplasm were extracted. Although there are methods for automated extraction of nucleus, we only focus on the classification task in this paper. We directly use the centroid of ground truth mask of nucleus to extract the image patches. 

Data augmentation is critical to improve the accuracy of CNNs and reducing overfitting. Because cervical cells are rotationally invariant, each cell image is performed $ N_r $ rotations with an angle step of $ \theta $ degree. Zero padding is also used to avoid region that lies outside of the image boundary. Considering that detecting the centroid of the nucleus may be inaccurate in practice, the centroid of each nucleus was randomly translated (by up to $d$ pixels) $ N_t $ times to obtain $ N_t $ coarse nucleus centers. Accordingly, $ N_t $ patches of size $ m \times m $ centered at these locations are extracted. These patches not only simulate inaccurate nucleus center detection, but also increase the amount of image samples for CNNs. Other data augmentation approaches such as scale and color transformations are not utilized, as both the size and intensity of the nucleus are essential features to classify abnormal cervical cells and normal ones. Note that the distribution of different types of cells in Herlev dataset is imbalance, so classifiers have a tendency to exhibit bias towards the majority classes. For example, the amount of abnormal cell images is larger than that of normal cell images in Herlev dataset. In order to balance the proportions of positive and negative samples, we apply a higher sampling proportion to the normal patches. For 7-class task, similar sampling methods are utilized for balancing.

\subsection{Convolutional Neural Network Architectures}
CNN is a deep learning model in which multiple stages of convolution, non-linearity and pooling layers are stacked, followed by more convolutional and fully-connected layers. In our experiments, we mainly explore four convolutional neural network models (AlexNet, GoogleNet, ResNet and DenseNet) which are shown in the green part in fig.\ref{fig_method}. The input of CNNs is image patch with five-channels which includes two channels of binary masks of the cervical nucleus and cytoplasm and three-channels of raw RGB image (left part in fig.\ref{fig_method}). To demonstrate the additive value of cell morphological features, raw RGB image is used as the only input of CNNs for performance comparison. The output layer of CNNs comprises of several neurons each corresponding to one class. In our case, there are 2 and 7 neurons in the output layer for 2-class and 7-class classification tasks, respectively. The backpropagation algorithm is used to minimize the classification error on the training dataset for optimization of weight parameters in CNNs. 

AlexNet \cite{krizhevsky2012imagenet}: AlexNet is the winner of ImageNet Large Scale Visual Recognition Challenge (ILSVRC) 2012 and has received extensive attention in computer vision. ImageNet dataset consists of 1.2 million $256 \times 256$ images belong to 1000 categories. AlexNet contains five convolution layers, three pooling layers, and three full-connected layers. AlexNet achieves 15.3\% top-5 classification error.

GoogleNet \cite{Szegedy2015going}: GoogleNet is more complex and deeper than AlexNet and is the winner of ImageNet ILSVRC 2014. GoogleNet introduces a new module named “inception”, which concatenates filters of different sizes and dimensions into a single new filter. Overall, GoogleNet has two convolution layers, two pooling layers, and nine “inception” layers. Each “inception” layer consists of six convolution layers and one pooling layer. GoogleNet obtains 6.67\% top-5 classification error on ImageNet dataset challenge.

ResNet \cite{He2016Deep}: ResNet is about 20 times deeper than AlexNet. ResNet utilizes shortcut connections to jump over some layers to avoid the problem of vanishing gradient. ResNet wins the ImageNet ILSVRC 2015, and have achieved impressive, record-breaking performance on many challenging image recognition, localization, and detection tasks \cite{He2016Deep}.

DenseNet \cite{Huang2017Densely}: Similar but different from ResNets, direct connections from any layer to all subsequent layers are introduced in DenseNets, which encourages feature reuse throughout the network. Moreover, the DenseNets can achieve state-of-the-art performances with substantially fewer parameters and less computation than ResNet.

\subsection{Transfer learning}
Transfer learning refers to the fine-tuning of deep learning models that are pre-trained on other large-scale image datasets. Due to limited cervical image data in this study, for each CNN architecture, pre-trained models trained on ImageNet dataset are used as the basis of our network, where the weights of the first convolution layer and last several task-specific full-connection layers are randomly initialized, and other network layers are transferred to the same locations. All of these layers are jointly trained on our cervical cell dataset. Note that only the first convolution layer and the last several task-specific full-connection layers are trained from scratch.

In testing, the random-view aggregation and multiple crop testing are used following the approach in \cite{Zhang2017DeepPap}.

\section{Experimental Methods}
\subsection{Data set}
The utilized cervical cell data is publicly available (\url{http://mdelab.aegean.gr/downloads}), which is collected at the Herlev University Hospital by a digital camera and microscope [16]. The image resolution is 0.201 $\mu m$ per pixel. The specimens are prepared via conventional Pap smear and Pap staining. There are 917 images in the Herlev dataset, where each image contains one cervical cell with its segmentation of nucleus and cytoplasm and the class label. In order to maximize the certainty of diagnosis, cervical images in Herlev dataset were diagnosed by two cyto-technicians and a doctor and were categorized into seven classes. These seven classes further belong to two categories: normal (class 1-3) and abnormal (class 4-7), as showed in Table \ref{herlevdataset}. For each cell image in the Herlev dataset, rotations and translations (up to 10 pixels) are performed to yields a relatively balanced data distribution. After augmentation, each class has roughly 12000 images. The RGB image patch size is set to $ m = 128 $ pixels to cover some cytoplasm region for most cells, and to contain most of the nucleus region for the largest one. Then segmentation masks of the nuclei and cytoplasm with the same size and location as RGB image patch are extracted. These image patches and masks are then up-sampled to a size of $ 256 \times 256 \times 3 $ and $ 256 \times 256 \times 2 $ pixels via nearest interpolation, in order to facilitate the transfer of pre-trained CNN model. The image patches and masks are concatenated to obtain five-channel dataset with a size of $ 256 \times 256 \times 5 $.

\begin{table}[!t]
\caption{The 917 cells (242 normal and 675 abnormal) from Herlev dataset.}
\label{herlevdataset}
\begin{tabular}{p{1cm}p{0.5cm}p{5.4cm}p{0.5cm}}
    \hline
     Category & Class & Cell type                                    & Num. \\
    \hline
    Normal   & 1     & Superficial squamous epithelial              & 74   \\
    Normal   & 2     & Intermediate squamous epithelial             & 70   \\
    Normal   & 3     & Columnar epithelial                          & 98   \\
    Abnormal & 4     & Mild squamous non-keratinizing dysplasia     & 182  \\
    Abnormal & 5     & Moderate squamous non-keratinizing dysplasia & 146  \\
    Abnormal & 6     & Severe squamous non-keratinizing dysplasia   & 197  \\
    Abnormal & 7     & Squamous cell carcinoma in situ intermediate & 150  \\
    \hline
\end{tabular}
    
\end{table}

\subsection{Network architectures and implementation}
In this study, two different inputs (i.e., raw RGB-channel dataset and five-channel dataset) are fed into four different CNN models, i.e., AlexNet, GoogleNet, ResNet-50 and DenseNet-121, and the classification performances for different tasks (2-class and 7-class problems) are compared. Note that there are deeper architectures for ResNet and DenseNet (e.g. ResNet-152, DenseNet-161). However, we found that ResNet-50 and DenseNet-121 have better performances than their deeper versions on our dataset. Here, the base CNN models (denoted as AlexNet-B, GoogleNet-B, ResNet-B, and DenseNet-B) are pre-trained on the ImageNet classification dataset. AlexNet-B contains three fully connection layer ($fc6$-$fc8$), and the number of neurons in last fully connection layer is determined by the number of output class. As shown in \cite{Zhang2017DeepPap}, reducing the number of neurons of $fc6$ and $fc7$ layer will tend to have slightly higher accuracy in this cervical dataset, and the neuron number of the $fc6$ and $fc7$ layer in our AlexNet model (denote as AlexNet-T) was set to be 1024-256. The AlexNet-T and the AlexNet-B share the same network, except for the first convolution layer and the three fully connection layers with weights of random initialized Gaussian distribution. For the other networks, weights of the first convolution layer and the last fully connection layer in our models (denote as GoogleNet-T, ResNet-T and DenseNet-T) are initialized with random Gaussian distribution, and the other initial weights are copied from the same location of GoogleNet-B, ResNet-B, and DenseNet-B, respectively. For each CNN model, we report classification results with different inputs and for 2-class and 7-class classification tasks: CNN-3C (three-channel RGB dataset input), CNN-5C (five-channel dataset input). Therefore, throughout this paper, we refer to a total of sixteen models. All these models are implemented on Caffe platform \cite{jia2014caffe}, using two Nvidia GeForce GTX 1080 Ti GPUs with a total memory of 22 GB.

\subsection{Training and Testing Protocols}
From each $256\times256$ training image patch and its mirrored version, a $227\times227$ sub-patch is randomly cropped for AlexNet-T, while a $224\times224$ sub-patch is randomly cropped for the other networks in this study. Stochastic Gradient Descent (SGD) is utilized to train the model for 30 epochs. The mini-batch sizes of training are 256, 32, 20 and 12 for AlexNet-T, GoogleNet-T, ResNet-T and DenseNet-T respectively. The base learning rates are 0.01, 0.005, 0.01 and 0.01 for for AlexNetT, GoogleNet-T, ResNet-T and DenseNet-T, respectively, and are decreased by a factor of 10 at every tenth epoch. Weight decay and momentum are set to be 0.0005 and 0.9 for AlexNet-T, and 0.0002 and 0.9 for the other networks.

\begin{table*}[!t]
    \centering
    \caption{Performance comparison of different models for 2-class classification task. Bold indicates the highest value in each column.}
    \label{Performance_2class}
    \begin{tabular}{p{3cm}p{2cm}p{2cm}p{2cm}p{2cm}}
           \hline
           Model & $AUC$  & $Acc(\%)$  &  $Sens(\%)$  &  $Spec(\%)$ \\
           \hline
           AlexNet-3C   & $0.962\pm0.008$  & $89.7\pm1.8$  &  $94.6\pm4.2$  &  $83.0\pm4.3$ \\ 
           AlexNet-5C   & $0.964\pm0.016$  & $91.5\pm2.8$  &  $96.5\pm2.9$  &  $84.7\pm4.1$ \\ 
           GoogleNet-3C & $0.979\pm0.005$  & $93.6\pm1.1$  &  $96.2\pm2.6$  &  $90.1\pm2.5$ \\ 
           GoogleNet-5C & $\textbf{0.984}\pm0.012$  & $\textbf{94.5}\pm2.8$  &  $\textbf{97.4}\pm2.7$  &  $\textbf{90.4}\pm3.1$ \\ 
           ResNet-3C    & $0.978\pm0.018$  & $92.3\pm2.6$  &  $94.8\pm3.3$  &  $89.1\pm6.3$ \\ 
           ResNet-5C    & $0.979\pm0.011$  & $92.1\pm2.0$  &  $97.3\pm2.8$  &  $85.2\pm4.3$ \\ 
           DenseNet-3C  & $0.970\pm0.013$  & $92.6\pm2.0$  &  $96.6\pm2.5$  &  $87.1\pm2.9$ \\ 
           DenseNet-5C  & $0.980\pm0.009$  & $93.3\pm2.0$  &  $95.6\pm2.8$  &  $90.0\pm3.6$ \\ 
           \hline
    \end{tabular}
    \end{table*}
    
\subsection{Evaluation Methods}
Previous methods using cross validation on the Herlev dataset do not guarantee patient-level splitting. While in real clinical practice, all the cells from a testing patient are unseen to the training set. Therefore, in this study, five-fold cross-validation is performed on patient-level. In each of the 5 iterations, 4 of 5 folds are used as the training set and the other one as validation set. We carefully ensure that cells of the same patient can only be in the training set or the validation set. Note that data augmentation is performed after the training/validation spitting of cell population. Final performances of models are obtained by averaging the results from 5 validation sets. The performance evaluation metrics include sensitivity ($Sens$), specificity ($Spec$), accuracy ($Acc$) and area under ROC curve ($AUC$), where $Sens$ indicates the proportion of correctly identified abnormal cells, $Spec$ is the proportion of correctly identified normal cells, and $Acc$ is the global percentage of correctly identified classified cell. The confusion matrix is used to show the classification performance of 7-class problem. The average accuarcy of classification of cervical cells is calculated by averaging the values on the main diagonal of confusion matrix.

\section{Results}
\begin{figure}[!t]
    \begin{center}
    \begin{tabular}{c}
    \includegraphics[width=8cm]{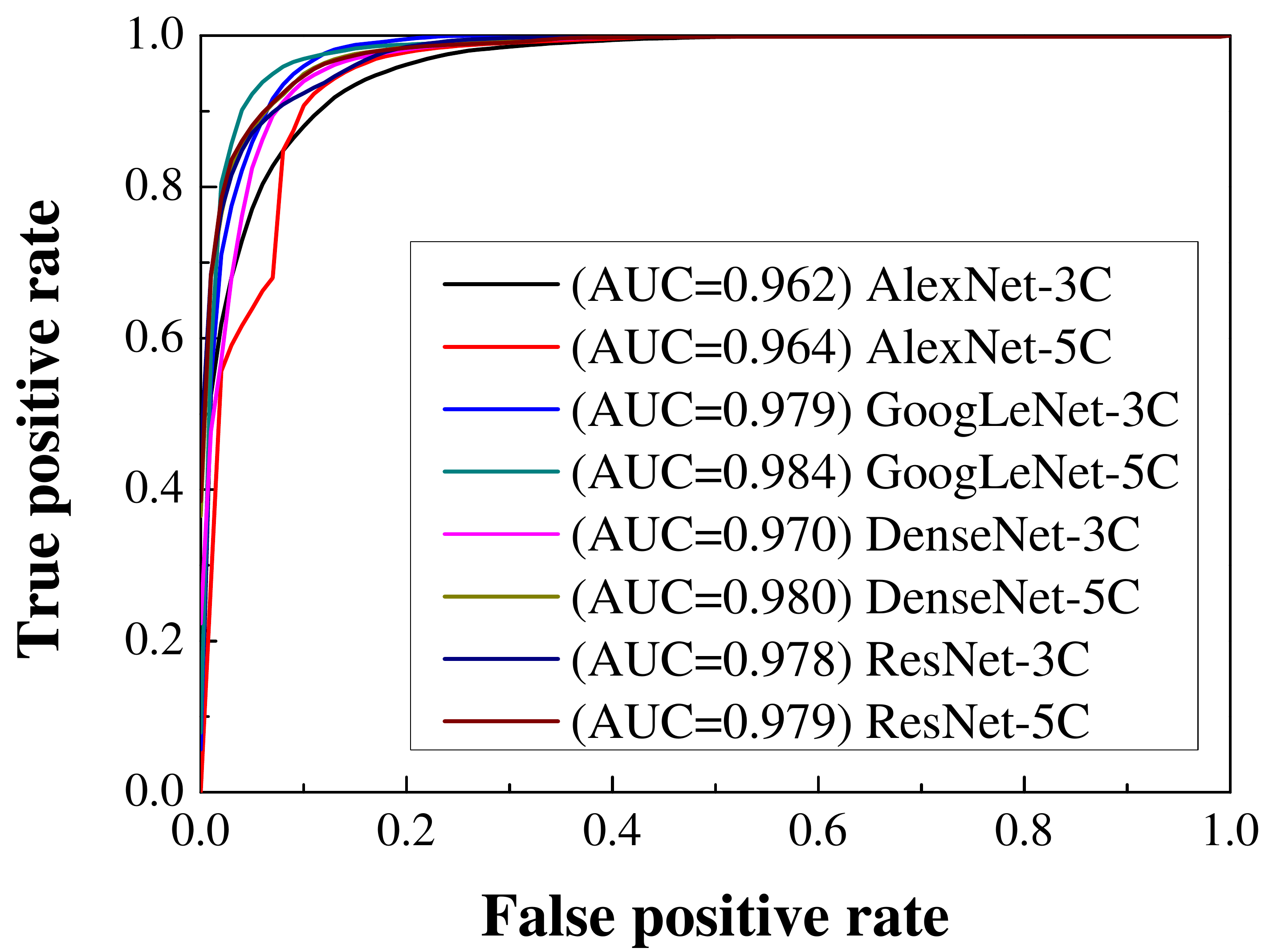}
    \end{tabular}
    \end{center}
    \caption[example] 
    { \label{Performace_ROC} 
    ROC curve comparison of different models for 2-class classification task.
 }
    \end{figure}

\begin{figure}[!t]
\begin{center}
    \begin{tabular}{c}
        \centering
    \includegraphics[width=8.8cm]{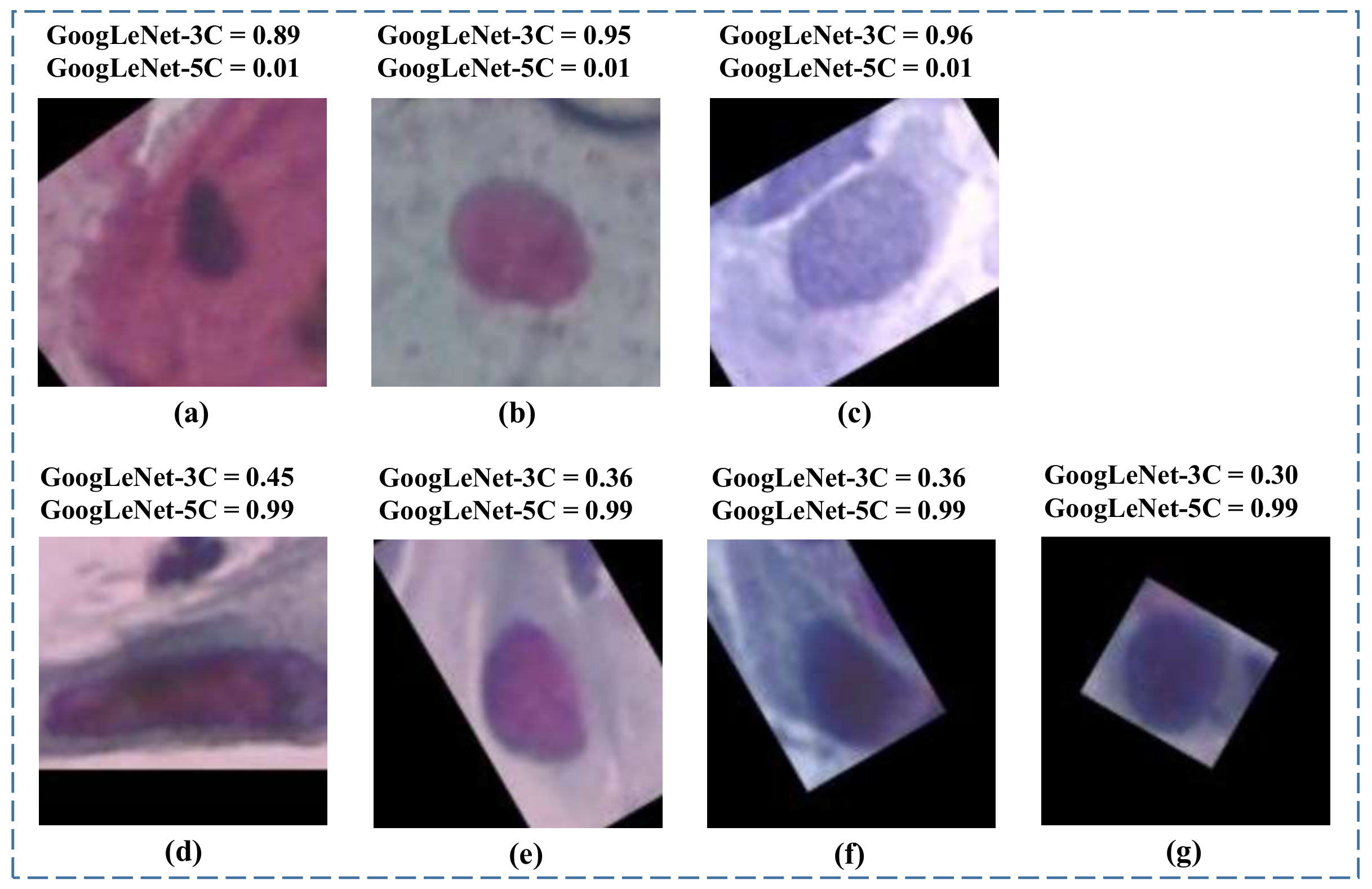}
    \end{tabular}
\end{center}
    \caption[example] 
    { \label{figure_example}
Examples of classified cervical cells by GoogleNet-3C and GoogleNet-5C. The ground truth labels of cells in the first row are Normal [(a) - (g) are superficial, intermediate and columnar] and the second row are Abnormal [(d) - (g) are mild dysplasia, moderate dysplasia, severe dysplasia and carcinoma]. Score = 1 corresponds a 100\% probability of representing an abnormal cell.
 }
\end{figure}

Table \ref{Performance_2class} and Fig. \ref{Performace_ROC} show the classification performances ($Sens$, $Spec$, $Acc$ and $AUC$) of CNN-5C model (AlexNet-5C, GoogleNet-5C, ResNet-5C and DenseNet-5C) in compared with CNN-3C model (AlexNet-3C, GoogleNet-3C, ResNet-3C and DenseNet-3C) for 2-class classification task. It can be seen that each of models with five-channel dataset as input outperforms its corresponding three-channel-input model in $AUC$ metrics. Among them, the model GoogleNet-5C obtains the best performance. The mean values of $Sens$, $Spec$, $Acc$ and $AUC$ for the model GoogleNet-5C are 97.4\%, 90.4\%, 94.5\% and 0.984, respectively. Some classification examples can be seen in Fig. \ref{figure_example}, where images misclassified by GoogleNet-3C can be correctly classified by GoogleNet-5C.

\begin{figure}[!t]
   \begin{center}
   \begin{tabular}{c}
   \includegraphics[width=8cm]{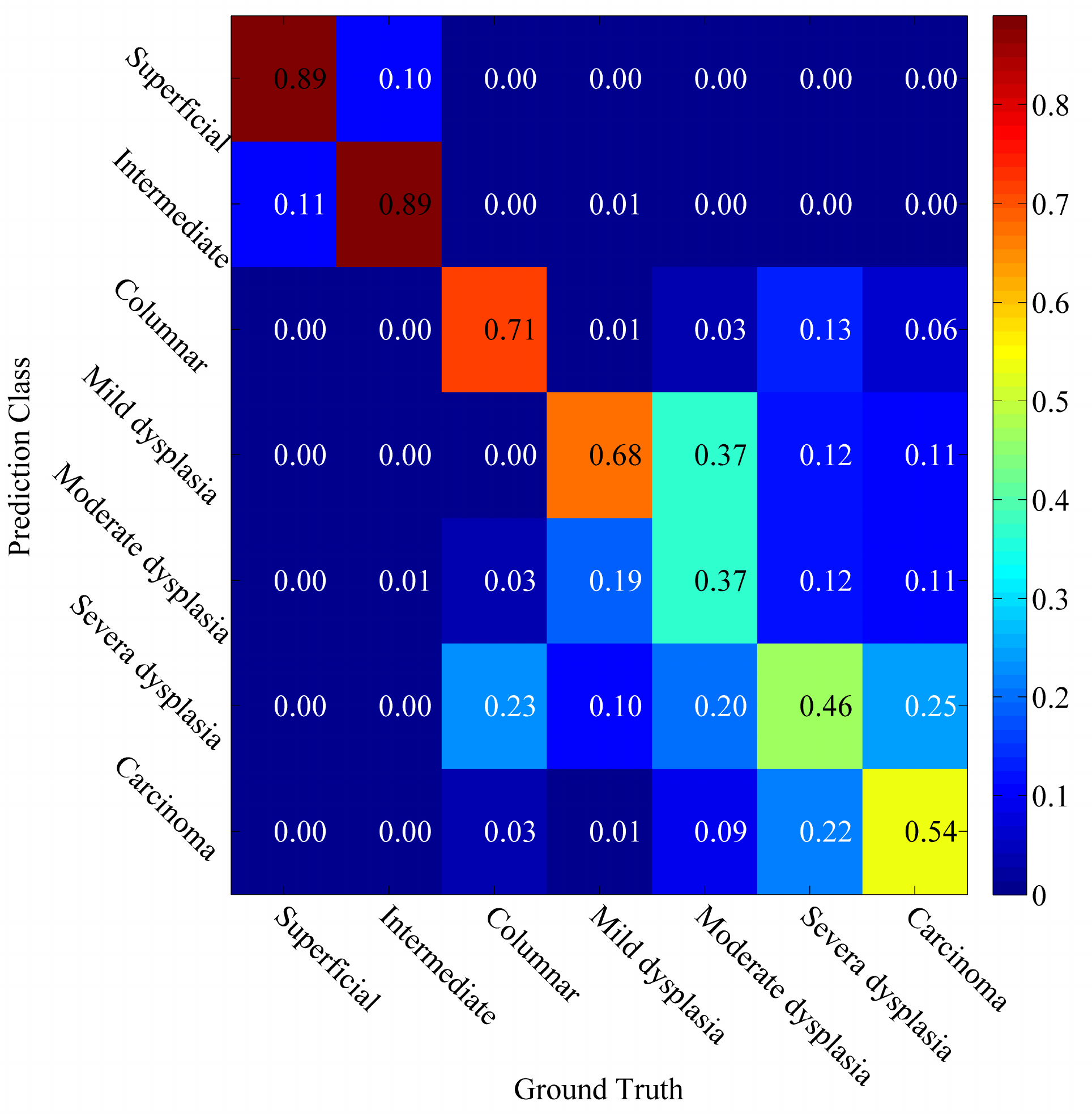}
   \end{tabular}
   \end{center}
   \caption[example] 
   { \label{figure_comfusionMatrix} 
   Confusion matrix of GoogleNet-5C for 7-class classification task.
}
   \end{figure}

Table 3 shows the classification performance of CNN-5C models and CNN-3C models for 7-class problem. CNN-5C models provide higher classification accuracy than CNN-3C models except for DenseNet. Among these models, GoogleNet-5C obtains the highest accuracy (64.5\%). Fig. \ref{figure_comfusionMatrix} shows the confusion matrix for the model of GoogleNet-5C, with the average accuracy (averaging the values on main diagonal of confusion matrix) of 64.8\%, which surpasses the previous non-deep-learning result of 61.1\% \cite{jantzen2005pap}.

\begin{table}[!t]
    \centering
    \caption{Accuracy comparison of different models for 7-class classification. Bold indicates the highest value in each column.}
    \begin{tabular}{p{3cm}p{2cm}}
           \hline
           Model & $Acc(\%)$ \\
           \hline
           AlexNet-3C   & $57.8\pm4.4$  \\ 
           AlexNet-5C   & $60.8\pm4.0$  \\ 
           GoogleNet-3C & $62.5\pm3.1$  \\ 
           \textbf{GoogleNet-5C} & $\textbf{64.5}\pm4.2$  \\ 
           ResNet-3C    & $60.8\pm3.7$  \\ 
           ResNet-5C    & $63.7\pm3.8$  \\ 
           DenseNet-3C  & $63.9\pm2.0$  \\ 
           DenseNet-5C  & $61.0\pm3.7$  \\ 
           \hline
    \end{tabular}
    \end{table}

\section{Discussion}
Combining traditional feature engineering with deep learning to improve classification performance is hot topic recently. Most of existing techniques perform classification by combining deep features learned by CNNs and other handcrafted features in traditional classifiers. In contrast, we directly fuse original RGB image of cervical cell with its nuclear and cytoplasmic binary masks as input for CNN, without relying on designing of other handcrafted features. For the 2-class classification problem, comparing with network without fusing morphological information of cervical nucleus and cytoplasm, the classification performances of appearance and morphology based CNNs are obviously higher. The GoogleNet-5C model has the highest classification accuracy, sensitivity, and specificity (94.5\%, 97.4\%, and 90.4\%). For the 7-class problem, the GoogleNet-5C model has the highest classification accuracy of 64.5\%. 

The networks deeper than GoogleNet, such as ResNet and DenseNet, do not provide higher performance on our task. This may due to the relatively small number of cells in Herlev cervical cell dataset used in this study, which may lead training on such complicate network to be overfit. For the 2-class problem, the best previous study obtained 96.8\% classification accuracy by optimizing the features derived from the manually segmented cytoplasm and nucleus \cite{marinakis2009pap}, which seems to be higher than the classification performance of the GoogleNet-5C model (94.5\%) in this study. This may be caused by the different cross-validation methods – our experiments are taken at patient-level data splitting, which is more closely related to the realistic clinical application. Also note that our method does not use any feature engineering, only raw RGB image and segmentation masks are used for CNN learning.

There is only one previous study (i.e., \cite{jantzen2005pap}) that reports detailed results of 7-class classification on Herlev dataset. In our study, GoogleNet-5C model obtains an average accuracy of 64.8\% for 7-class problem, which is 3.7\% higher than that of 61.1\% in [16]. As shown in the confusion matrix in Fig.\ref{figure_comfusionMatrix}, some cells are easier to be classified, while some are harder; superficial and intermediate cells are classified with the highest accuracy. Similar to results in [16], some columnar cells are wrongly classified as severe dysplasia cells because severe dysplasia cells have similar characteristics in appearance and morphology with columnar cells (e.g., dark nuclei and small-sized cytoplasm); The fine-grained classification of abnormal cells (i.e., mild dysplasia, moderate dysplasia, severe dysplasia and carcinoma) remains very challenging. In general, such a task is hard even for cyto-pathologists but highly desirable and with significantly clinical value; The most difficult case is moderate abnormal cell with correct classification rate of only 37\%.

The segmentation of nucleus and cytoplasm are pre-required for applying our method. This is directly obtained from the ground truth segmentation in this paper. Note that screening of abnormal of cells in practice using the proposed method requires automated segmentation of the nucleus and cytoplasm. This task may be achieved by using fully convolutional networks (e.g., U-Net) \cite{Ronneberger2015U-Net} or specifically designed algorithm \cite{zhang2017combining} for semantic segmentation of cervical cells. The effects of automated segmentation on classification performance still need to be analyzed in future study. Nevertheless, the current results demonstrate that fine-grained classification of cervical cells into different abnormal levels remains to be very challenging even with accurate cell segmentation available. 

\section{Conclusion}
This paper proposes an appearance and morphology based convolutional neural network method for cervical cell classification. Unlike the previous CNN-based method which only uses raw image data as network input, our method combines the raw image data with segmentation masks of the nucleus and cytoplasm as network input. Our method consists of extracting cell image/mask patches coarsely centered on the nucleus, transferring features from another pre-trained model into a new model for fine-tuning on the cervical cell image dataset, and forming the final network output. State-of-the-art CNN networks including AlexNet, GoogleNet, ResNet and DenseNet are trained for performance comparison. The results show that the combination of raw RGB data with segmentation masks of nuclei and cytoplasm as network input can provide higher performance in the fine-grained classification of cervical cells, which has important clinical importance in early diagnosis of cervical cells dysplastic changes.


\ifCLASSOPTIONcaptionsoff
  \newpage
\fi

\bibliography{ref_cervical}   

\begin{thebibliography}{10}
\providecommand{\url}[1]{#1}
\csname url@samestyle\endcsname
\providecommand{\newblock}{\relax}
\providecommand{\bibinfo}[2]{#2}
\providecommand{\BIBentrySTDinterwordspacing}{\spaceskip=0pt\relax}
\providecommand{\BIBentryALTinterwordstretchfactor}{4}
\providecommand{\BIBentryALTinterwordspacing}{\spaceskip=\fontdimen2\font plus
\BIBentryALTinterwordstretchfactor\fontdimen3\font minus
  \fontdimen4\font\relax}
\providecommand{\BIBforeignlanguage}[2]{{%
\expandafter\ifx\csname l@#1\endcsname\relax
\typeout{** WARNING: IEEEtran.bst: No hyphenation pattern has been}%
\typeout{** loaded for the language `#1'. Using the pattern for}%
\typeout{** the default language instead.}%
\else
\language=\csname l@#1\endcsname
\fi
#2}}
\providecommand{\BIBdecl}{\relax}
\BIBdecl

\bibitem{Jemal2011Global}
A.~Jemal, F.~Bray, M.~M. Center, J.~Ferlay, E.~Ward, and D.~Forman, ``{Global
  Cancer Statistics},'' \emph{{CA-A Cancer Journal for Clinicians}}, vol.~{61},
  no.~{2}, pp. {69--90}, {2011}.

\bibitem{Gadducci2011smoking}
A.~Gadducci, C.~Barsotti, S.~Cosio, L.~Domenici, and A.~R. Genazzani,
  ``{Smoking habit, immune suppression, oral contraceptive use, and hormone
  replacement therapy use and cervical carcinogenesis: a review of the
  literature},'' \emph{{Gynecological Endocrinology}}, vol.~{27}, no.~{8}, pp.
  {597--604}, {2011}.

\bibitem{saslow2012american}
D.~Saslow, D.~Solomon, H.~W. Lawson, M.~Killackey, S.~L. Kulasingam, J.~Cain,
  F.~A. Garcia, A.~T. Moriarty, A.~G. Waxman, D.~C. Wilbur \emph{et~al.},
  ``American cancer society, american society for colposcopy and cervical
  pathology, and american society for clinical pathology screening guidelines
  for the prevention and early detection of cervical cancer,'' \emph{CA: A
  Cancer Journal for Clinicians}, vol.~62, no.~3, pp. 147--172, 2012.

\bibitem{birdsong1996automated}
G.~G. Birdsong, ``Automated screening of cervical cytology specimens,''
  \emph{Human Pathology}, vol.~27, no.~5, pp. 468--481, 1996.

\bibitem{zhang2014automation}
L.~Zhang, H.~Kong, C.~T. Chin, S.~Liu, X.~Fan, T.~Wang, and S.~Chen,
  ``Automation-assisted cervical cancer screening in manual liquid-based
  cytology with hematoxylin and eosin staining,'' \emph{Cytom. Part A},
  vol.~85, no.~3, pp. 214--230, 2014.

\bibitem{bengtsson2014screening}
E.~Bengtsson and P.~Malm, ``Screening for cervical cancer using automated
  analysis of pap-smears,'' \emph{Comput. Math. Method Med.}, vol. 2014, pp.
  1--12, 2014.

\bibitem{demay2007practical}
R.~M. DeMay, \emph{Practical principles of cytopathology}, revised
  edition~ed.\hskip 1em plus 0.5em minus 0.4em\relax Chicago: IL: American
  Society for Clinical Pathology Press, 2007.

\bibitem{canavan2000cervical}
T.~Canavan and N.~Doshi, ``{Cervical cancer},'' \emph{{AMERICAN FAMILY
  PHYSICIAN}}, vol.~{61}, no.~{5}, pp. {1369--1376}, {2000}.

\bibitem{gao2016hep}
Z.~Gao, L.~Wang, L.~Zhou, and J.~Zhang, ``Hep-2 cell image classification with
  deep convolutional neural networks,'' \emph{IEEE Journal of Biomedical and
  Health Informatics}, 2016.

\bibitem{zhang2014segmentation}
L.~Zhang, H.~Kong, C.~T. Chin, S.~Liu, Z.~Chen, T.~Wang, and S.~Chen,
  ``Segmentation of cytoplasm and nuclei of abnormal cells in cervical cytology
  using global and local graph cuts,'' \emph{Comput. Med. Imaging Graph.},
  vol.~38, no.~5, pp. 369--380, 2014.

\bibitem{chankong2014automatic}
T.~Chankong, N.~Theera-Umpon, and S.~Auephanwiriyakul, ``Automatic cervical
  cell segmentation and classification in pap smears,'' \emph{Comput. Meth.
  Programs Biomed.}, vol. 113, no.~2, pp. 539--556, 2014.

\bibitem{plissiti2012overlapping}
M.~E. Plissiti and C.~Nikou, ``Overlapping cell nuclei segmentation using a
  spatially adaptive active physical model,'' \emph{IEEE Trans. Image
  Process.}, vol.~21, no.~11, pp. 4568--4580, 2012.

\bibitem{guo2012discriminative}
Y.~Guo, G.~Zhao, and M.~Pietik{\"a}Inen, ``Discriminative features for texture
  description,'' \emph{Pattern Recognition}, vol.~45, no.~10, pp. 3834--3843,
  2012.

\bibitem{nanni2010local}
L.~Nanni, A.~Lumini, and S.~Brahnam, ``Local binary patterns variants as
  texture descriptors for medical image analysis,'' \emph{Artificial
  Intelligence in Medicine}, vol.~49, no.~2, pp. 117--125, 2010.

\bibitem{marinakis2008particle}
Y.~Marinakis, M.~Marinaki, and G.~Dounias, ``Particle swarm optimization for
  pap-smear diagnosis,'' \emph{Expert Systems with Applications}, vol.~35,
  no.~4, pp. 1645--1656, 2008.

\bibitem{jantzen2005pap}
J.~Jantzen, J.~Norup, G.~Dounias, and B.~Bjerregaard, ``Pap-smear benchmark
  data for pattern classification,'' \emph{Nature inspired Smart Information
  Systems (NiSIS 2005)}, pp. 1--9, 2005.

\bibitem{van2002automated}
J.~van~der Laak, A.~Siebers, V.~Cuijpers, M.~Pahlplatz, P.~de~Wilde, and
  A.~Hanselaar, ``{Automated identification of diploid reference cells in
  cervical smears using image analysis},'' \emph{{Cytometry}}, vol.~{47},
  no.~{4}, pp. {256--264}, {2002}.

\bibitem{tucker1998Interval}
J.~Tucker, K.~Rodenacker, U.~Juetting, P.~Nickolls, K.~Watts, and G.~Burger,
  ``{Interval-coded texture features for artifact rejection in automated
  cervical cytology},'' \emph{{Cytometry}}, vol.~{9}, no.~{5}, pp. {418--425},
  {1988}.

\bibitem{plissiti2011combining}
M.~E. Plissiti, C.~Nikou, and A.~Charchanti, ``{Combining shape, texture and
  intensity features for cell nuclei extraction in Pap smear images},''
  \emph{{Pattern Recognition Letters}}, vol.~{32}, no.~{6}, pp. {838--853},
  {2011}.

\bibitem{Jusman2014Intelligent}
Y.~Jusman, S.~C. Ng, and N.~A. Abu~Osman, ``{Intelligent Screening Systems for
  Cervical Cancer},'' \emph{{Scientific World Journal}}, {2014}.

\bibitem{bora2017automated}
K.~Bora, M.~Chowdhury, L.~B. Mahanta, M.~K. Kundu, and A.~K. Das, ``Automated
  classification of pap smear images to detect cervical dysplasia,''
  \emph{Computer Methods and Programs in Biomedicine}, vol. 138, pp. 31--47,
  2017.

\bibitem{lecun2015deep}
Y.~LeCun, Y.~Bengio, and G.~Hinton, ``Deep learning,'' \emph{Nature}, vol. 521,
  no. 7553, pp. 436--444, 2015.

\bibitem{lecun1989backpropagation}
Y.~LeCun, B.~Boser, J.~S. Denker, D.~Henderson, R.~E. Howard, W.~Hubbard, and
  L.~D. Jackel, ``Backpropagation applied to handwritten zip code
  recognition,'' \emph{Neural Computation}, vol.~1, no.~4, pp. 541--551, 1989.

\bibitem{krizhevsky2012imagenet}
A.~Krizhevsky, I.~Sutskever, and G.~E. Hinton, ``Image{N}et classification with
  deep convolutional neural networks,'' in \emph{Advances in Neural Information
  Processing Systems}, 2012, pp. 1097--1105.

\bibitem{Szegedy2015going}
C.~Szegedy, W.~Liu, Y.~Jia, P.~Sermanet, S.~Reed, D.~Anguelov, D.~Erhan,
  V.~Vanhoucke, and A.~Rabinovich, ``{Going Deeper with Convolutions},'' in
  \emph{{IEEE Conference on Computer Vision and Pattern Recognition
  (CVPR)}}.\hskip 1em plus 0.5em minus 0.4em\relax {IEEE}, {2015}, pp. {1--9}.

\bibitem{He2016Deep}
K.~He, X.~Zhang, S.~Ren, and J.~Sun, ``{Deep Residual Learning for Image
  Recognition},'' in \emph{{IEEE Conference on Computer Vision and Pattern
  Recognition (CVPR)}}, ser. {IEEE Conference on Computer Vision and Pattern
  Recognition}.\hskip 1em plus 0.5em minus 0.4em\relax {IEEE}, {2016}, pp.
  {770--778}.

\bibitem{Huang2017Densely}
G.~Huang, Z.~Liu, L.~van~der Maaten, and K.~Q. Weinberger, ``{Densely Connected
  Convolutional Networks},'' in \emph{{IEEE Conference on Computer Vision and
  Pattern Recognition (CVPR 2017)}}, ser. {IEEE Conference on Computer Vision
  and Pattern Recognition}.\hskip 1em plus 0.5em minus 0.4em\relax {IEEE},
  {2017}, pp. {2261--2269}.

\bibitem{yosinski2014transferable}
J.~Yosinski, J.~Clune, Y.~Bengio, and H.~Lipson, ``How transferable are
  features in deep neural networks?'' in \emph{Advances in Neural Information
  Processing Systems}, 2014, pp. 3320--3328.

\bibitem{shin2016deep}
H.-C. Shin, H.~R. Roth, M.~Gao, L.~Lu, Z.~Xu, I.~Nogues, J.~Yao, D.~Mollura,
  and R.~M. Summers, ``Deep convolutional neural networks for computer-aided
  detection: {CNN} architectures, dataset characteristics and transfer
  learning,'' \emph{IEEE Transactions on Medical Imaging}, vol.~35, no.~5, pp.
  1285--1298, 2016.

\bibitem{wang2017chestx}
X.~Wang, Y.~Peng, L.~Lu, Z.~Lu, M.~Bagheri, and R.~M. Summers, ``Chestx-ray8:
  Hospital-scale chest x-ray database and benchmarks on weakly-supervised
  classification and localization of common thorax diseases,'' in
  \emph{Computer Vision and Pattern Recognition (CVPR), 2017 IEEE Conference
  on}.\hskip 1em plus 0.5em minus 0.4em\relax IEEE, 2017, pp. 3462--3471.

\bibitem{de2018clinically}
J.~De~Fauw, J.~R. Ledsam, B.~Romera-Paredes, S.~Nikolov, N.~Tomasev,
  S.~Blackwell, H.~Askham, X.~Glorot, B.~O’Donoghue, D.~Visentin
  \emph{et~al.}, ``Clinically applicable deep learning for diagnosis and
  referral in retinal disease,'' \emph{Nature Medicine}, vol.~24, no.~9, p.
  1342, 2018.

\bibitem{yan2018deep}
K.~Yan, X.~Wang, L.~Lu, L.~Zhang, A.~Harrison, M.~Bagheri, and R.~M. Summers,
  ``Deep lesion graphs in the wild: relationship learning and organization of
  significant radiology image findings in a diverse large-scale lesion
  database,'' in \emph{IEEE CVPR}, 2018.

\bibitem{zhang2018convolutional}
L.~Zhang, L.~Lu, R.~M. Summers, E.~Kebebew, and J.~Yao, ``Convolutional
  invasion and expansion networks for tumor growth prediction,'' \emph{IEEE
  Transactions on Medical Imaging}, vol.~37, no.~2, pp. 638--648, 2018.

\bibitem{NANNI2018ensemble}
\BIBentryALTinterwordspacing
L.~Nanni, S.~Ghidoni, and S.~Brahnam, ``Ensemble of convolutional neural
  networks for bioimage classification,'' \emph{Applied Computing and
  Informatics}, 2018. [Online]. Available:
  \url{http://www.sciencedirect.com/science/article/pii/S2210832718301388}
\BIBentrySTDinterwordspacing

\bibitem{Zhang2017DeepPap}
L.~Zhang, L.~Lu, I.~Nogues, R.~M. Summers, S.~Liu, and J.~Yao, ``Deeppap: Deep
  convolutional networks for cervical cell classification,'' \emph{IEEE Journal
  of Biomedical and Health Informatics}, vol.~21, no.~6, pp. 1633--1643, Nov
  2017.

\bibitem{jia2014caffe}
Y.~Jia, E.~Shelhamer, J.~Donahue, S.~Karayev, J.~Long, R.~Girshick,
  S.~Guadarrama, and T.~Darrell, ``Caffe: Convolutional architecture for fast
  feature embedding,'' in \emph{Proceedings of the 22nd ACM International
  Conference on Multimedia}.\hskip 1em plus 0.5em minus 0.4em\relax ACM, 2014,
  pp. 675--678.

\bibitem{marinakis2009pap}
Y.~Marinakis, G.~Dounias, and J.~Jantzen, ``Pap smear diagnosis using a hybrid
  intelligent scheme focusing on genetic algorithm based feature selection and
  nearest neighbor classification,'' \emph{Computers in Biology and Medicine},
  vol.~39, no.~1, pp. 69--78, 2009.

\bibitem{Ronneberger2015U-Net}
O.~Ronneberger, P.~Fischer, and T.~Brox, ``{U-Net: Convolutional Networks for
  Biomedical Image Segmentation},'' in \emph{{Medical Image Computing and
  Computer-Assisted Intervention (MICCAI), PT III}}, vol. {9351}.\hskip 1em
  plus 0.5em minus 0.4em\relax {Springer}, {2015}, pp. {234--241}.

\bibitem{zhang2017combining}
L.~Zhang, M.~Sonka, L.~Lu, R.~M. Summers, and J.~Yao, ``Combining fully
  convolutional networks and graph-based approach for automated segmentation of
  cervical cell nuclei,'' in \emph{2017 IEEE 14th International Symposium on
  Biomedical Imaging}, 2017.

\end{thebibliography}
\bibliographystyle{IEEEtran}   

\end{document}